\documentclass[sigconf]{acmart}
\AtBeginDocument{%
  }

\copyrightyear{2025}
\acmYear{2025}
\setcopyright{cc}
\setcctype{by}
\acmConference[SIGSPATIAL '25]{The 33rd ACM International Conference on Advances in Geographic Information Systems}{November 3--6, 2025}{Minneapolis, MN, USA}
\acmBooktitle{The 33rd ACM International Conference on Advances in Geographic Information Systems (SIGSPATIAL '25), November 3--6, 2025, Minneapolis, MN, USA}
\acmDOI{10.1145/3748636.3762783}
\acmISBN{979-8-4007-2086-4/2025/11}

\usepackage{listings}
\usepackage{multirow}
\usepackage{multicol}
\usepackage{booktabs}
\usepackage{graphicx}  
\usepackage{array}

\definecolor{codegreen}{rgb}{0,0.6,0}
\definecolor{codegray}{rgb}{0.5,0.5,0.5}
\definecolor{codepurple}{rgb}{0.58,0,0.82}
\definecolor{backcolour}{rgb}{0.95,0.95,0.92}

\lstdefinestyle{mystyle}{
    backgroundcolor=\color{backcolour},   
    commentstyle=\color{codegreen},
    keywordstyle=\color{magenta},
    numberstyle=\tiny\color{codegray},
    stringstyle=\color{codepurple},
    basicstyle=\ttfamily\footnotesize,
    breakatwhitespace=false,         
    breaklines=true,                 
    captionpos=b,                    
    keepspaces=true,                 
    numbers=none,                    
    numbersep=3pt,                  
    showspaces=false,                
    showstringspaces=false,
    showtabs=false,                  
    tabsize=2
}

\begin{document}

\title{OBSR: Open Benchmark for Spatial Representations}


\author{Julia Moska}
\orcid{0009-0003-8581-1098}
\authornote{Equal contribution to the benchmark}
\author{Oleksii Furman}
\authornotemark[1]
\orcid{0009-0001-2184-3096}
\affiliation{%
  \institution{Wrocław University of Science and Technology}
  \department{Department of Artificial Intelligence \\ Kraina.AI}
  \city{Wrocław}
  \country{Poland}
}
\email{julia.moska@pwr.edu.pl}

\author{Kacper Kozaczko}
\authornotemark[1]
\orcid{0009-0002-8668-4652}
\author{Szymon Leszkiewicz}
\authornotemark[1]
\orcid{0009-0001-5785-4184}
\author{Jakub Polczyk}
\authornotemark[1]
\orcid{0009-0002-5987-5146}
\affiliation{%
  \institution{Kraina.AI}
  \city{Wrocław}
  \country{Poland}
}

\author{Piotr Gramacki}
\orcid{0000-0002-4587-5586}
\author{Piotr Szymański}
\email{piotr.szymanski@pwr.edu.pl}
\orcid{0000-0002-7733-3239}
\affiliation{%
  \institution{Wrocław University of Science and Technology}
  \department{Department of Artificial Intelligence \\ Kraina.AI}
  \city{Wrocław}
  \country{Poland}
}

\renewcommand{\shortauthors}{Moska et al.}

\begin{abstract}
GeoAI is evolving rapidly, fueled by diverse geospatial datasets like traffic patterns, environmental data, and crowdsourced OpenStreetMap (OSM) information. While sophisticated AI models are being developed, existing benchmarks are often concentrated on single tasks and restricted to a single modality. As such, progress in GeoAI is limited by the lack of a standardized, multi-task, modality-agnostic benchmark for their systematic evaluation. This paper introduces a novel benchmark designed to assess the performance, accuracy, and efficiency of geospatial embedders. Our benchmark is modality-agnostic and comprises 7 distinct datasets from diverse cities across three continents, ensuring generalizability and mitigating demographic biases. It allows for the evaluation of GeoAI embedders on various phenomena that exhibit underlying geographic processes. Furthermore, we establish a simple and intuitive task-oriented model baselines, providing a crucial reference point for comparing more complex solutions. 
\end{abstract}




\maketitle

\section{Introduction}
The rapid advancements in artificial intelligence (AI) have transformed many industries, thanks in large part to the establishment of robust benchmarking systems that facilitate performance evaluation and comparison. In the realm of geospatial analysis, the emergence of sophisticated AI models has highlighted the critical need for specialized benchmarks tailored to this unique data landscape. This paper introduces a geospatial benchmark designed to assess the performance, accuracy, and efficiency of AI-driven geospatial tools and algorithms. By using standardized metrics and protocols, this benchmark will not only enable objective comparisons among various AI solutions but also encourage innovations that leverage geospatial data. 

Geospatial data, essential for these applications, must either represent geometries with coordinates or be geocodable to specific regions or locations. Such data can encompass various forms, including text (e.g., tags, reviews, social media posts, documents), images (e.g., satellite multispectral and hyperspectral imagery, aerial photos, street view images, videos), and numerical values (e.g., weather readings, population statistics, traffic data, point-of-interest ratings). As such most of the existing benchmarks have been limited to a single optimization task ex. location encoding, or modality such ex. satelite imagery. What is however not available is a multi-task benchmark that is agnostic of the modality, featuring an intuitive baseline to compare to - based on basic map-sourced information only. 

The contributions of this paper are following:
\begin{enumerate}
    \item review of established open datasets for popular tasks in geospatial research community,
    \item discussion on what makes a geospatial benchmark work and data processing needed for benchmark creation,
    \item multi-task, modality agnostic benchmark that allows evaluation of GeoAI models' capability of inference in terms of multiple phenomena that exhibit an underlying geographic process and are well cited in scientific literature,
    \item natural, simple and intuitive baseline that the authors of large geospatial models can compare to.
\end{enumerate}

We use the \textsc{SRAI} \cite{SRAI} library, which provides a critical resource by focusing on publicly available datasets with clearly defined vector geometries (points, lines, polygons). This focus enhances reproducibility, simplifies usage, and democratizes access to geospatial data, enabling users without extensive GIS expertise to engage with and benefit from geospatial analysis. This library is a natural home for the kind of benchmark we are proposing. Benchmark code is available in public GitHub repository: \url{https://github.com/kraina-ai/srai}, where we provide a reproducible experimental code. 

\section{Related work}
In this Section we provide an overview of best practices in benchmarking across different domains and existing geospatial benchmarks. Based on that, we define requirements for a high quality geospatial benchmark and show which gaps we aim to fill with our proposed OBSR benchmark.

\subsection{Non-Geospatial Benchmarks}

The development of standardized datasets and benchmarks has been fundamental to recent breakthroughs in machine learning and artificial intelligence. Well-established, universally adopted evaluation frameworks have enabled significant progress across domains including Computer Vision and Natural Language Processing. As models demonstrate improved performance on existing datasets, this success drives the creation of more challenging evaluation tasks, creating a virtuous cycle of advancement. With models becoming increasingly capable of handling multiple tasks simultaneously, the focus shifted toward comprehensive multi-task evaluation frameworks. Benchmarks such as GLUE\cite{wang2019gluemultitaskbenchmarkanalysis} and SuperGLUE\cite{sarlin2020supergluelearningfeaturematching} established the standard for systematic language model evaluation, an approach subsequently adapted for other languages through initiatives like LEPISZCZE\cite{augustyniak2022waydesigningcompilinglepiszcze} for Polish. Beyond merely aggregating datasets, LEPISZCZE introduced crucial infrastructure improvements including unified access APIs and standardized train/test splits, facilitating reproducible research and rigorous model comparisons. The MTEB\cite{muennighoff2023mtebmassivetextembedding} and MMTEB\cite{enevoldsen2025mmtebmassivemultilingualtext} benchmarks further demonstrate the potential of comprehensive evaluation frameworks, providing unified interfaces for diverse text embedding tasks while maintaining public leaderboards that enable practitioners to identify optimal models for specific applications.

Computer vision benchmarks illustrate similar design evolution: from ImageNet's\cite{imagenet} emphasis on scale and categorical diversity (14 million images across 20,000 categories) to MS COCO's\cite{ms_coco} focus on contextual complexity and task variety (object detection, segmentation, and captioning in natural scenes).

Effective benchmark design principles, demonstrated across NLP and computer vision domains, emphasize domain diversity and real-world contextual complexity that mirrors actual deployment scenarios rather than simplified academic tasks. These design patterns ensure that benchmarks drive meaningful progress by exposing the gap between controlled experimental performance and practical application requirements.

\subsection{Geospatial benchmarks}

Geospatial community is also developing benchmarks to evaluate models in rapidly developing GeoAI domain, which is starting to develop Geospatial Foundation Models (GeoFM). This leads to new challenges, raised by \citet{mai_towards_2025}, one of which is the need for a unified set of geospatial tasks to evaluate those GeoFMs on. This need for more geospatial benchmarks was highlighted by the IBM/NASA team that built Prithvi - the first open source GeoFM \cite{jakubik2023foundation}, reiterated by scientists working on Urban Foundation Models (UFM) \cite{zhang2025urbangeneralintelligencereview}. Scientists who advocate for coherent experimental AI research based on scientific principles published a document mentioning the need for new benchmarks recently as well \cite{li2024geoai}.

While reviewing the literature in search of GeoAI benchmarks one can immediately see that existing benchmarks are limited to a single modality and often to a single task or a family of very close tasks. 

Computer Vision related geospatial benchmarks are usually dedicated to models that work on remote sensing imagery either satellite-captured or obtained through orthophotography. The authors of TorchGeo\cite{stewart2022torchgeodeeplearninggeospatial} noted that many datasets used in geospatial imagery exhibit caveats and errors which require data processing, which was often performed differently by various authors thus reporting results that were non-comparable. As a result, the authors provide a library that, among others, standardizes the dataset preparation and allows comparable benchmarking.

Following the success of ImageNet, the GeoImageNet\cite{li2023geoimagenet} has been proposed. This dataset focuses on benchmarking the models with respect to their ability to detect various qualities of terrain based on satellite imagery. Another recent satellite imagery benchmark was dedicated to measuring the ability of satellite image segmentation \cite{wang2023samrs}. 

Earth monitoring benchmarking has been another area of recent advances, which include: Geo-BENCH\cite{lacoste2023geobenchfoundationmodelsearth}, EarthNets\cite{xiong2024earthnetsempoweringaiearth}, or SustainBench\cite{yeh_sustainbench_2021}. These benchmarks are in use by most teams that deliver GeoFMs for satellite imagery at global scale.

\citet{shi2023opentransmindnewbaselinebenchmark} propose the first benchmark targeted for models that provide inference in terms of estimating various traffic tasks from different scale imagery, including pedestrian detection and vehicle flow counting.

Many problems in GeoAI are related to either textual inference in terms of sequences of spatial data points, or treating spatial sequences as token sequences and applying approaches from Natural Language Processing domain. Several benchmarks have been published in relation to these cases. \citet{xie2024travelplanner} introduced a benchmark to evaluate a language model's capability to generate a coherent travel itinerary. \citet{aghzal2025largelanguagemodelsgood} is a benchmark of an LLM's capability to generate sequences of spatial points. \citet{Yang15042025} proposes a benchmark to evaluate the capability of language models to perform spatial reasoning and comprehension.

Finally, TorchSpatial\cite{wu2025torchspatiallocationencodingframework} recently proposed an image-based benchmark for evaluating the efficiency of location embedding.

To summarise, there are already some high-quality benchmarks in the geospatial domain, however there is a crucial gap that needs to be addressed. A vast majority of existing benchmarks focus on remote sensing data and tasks. While they cover some vital applications, their applicability to urban-related tasks is limited. Therefore, there is a need for a benchmark designed with other modalities in mind, such as vector data from OSM.

\section{Benchmark}

In this Section we describe the OBSR benchmark in detail. We first define a set of criteria that are essential for high quality geospatial benchmark from two perspectives: (i) general benchmark design and (ii) geospatial specific requirements for urban-related applications. We then follow with a detailed description of tasks which we include in our benchmark. Finally, we present which of the highly esteemed datasets we used to compile our benchmark and how they were processed to create a coherent benchmark.

\subsection{What is a good geospatial benchmark?}

Based on related works and specific geospatial domain characteristics, we extracted a set of requirements which are crucial for a high-quality benchmark:

\begin{itemize}
    \item Wide range of datasets covering various tasks in the domain. We achieve this by combining well established datasets and unifying access to them using a single library and data format. We also select different tasks and types of geometries.

    \item Reproducibility is provided by clearly defined train/test splits. We provide both raw data and predefined splits with preprocessing already applied. This makes it easy to rerun experiments and test new models on the same instances of a given problem.

    \item Granularity across tasks aims to adapt to different types of spatial dependencies. We achieve this by using different H3\footnote{https://www.uber.com/en-PL/blog/h3/} resolutions to aggregate data and create instances for datasets in our benchmark. 

    \item Accessibility of data and evaluation code. We host all raw and processed data in publicly available HuggingFace repository and provide an extension to the SRAI library, which handles loading and preprocessing datasets. This integration facilitates seamless use of embedding models implemented within the SRAI framework. To ensure broader accessibility, we also offer straightforward mechanisms for evaluating models independently of the SRAI ecosystem.

\end{itemize}

\subsection{Datasets}

\begin{table*}[t]
\caption{Summary of datasets included in benchmark. Includes spatio-temporal coverage and number of instances in splits.}
\label{tab:splits}
\begin{tabular}{l|cc|rr|rr|rr}
\toprule
                            & &  & \multicolumn{2}{c}{res8} & \multicolumn{2}{c}{res9} & \multicolumn{2}{c}{res10} \\
                            & Spatial coverage & Temporal Coverage  & train       & test       & train       & test       & train       & test        \\ \midrule
Airbnb  & 6 Cities (EU, US, AU) & 2022-2023                   & 6648        & 5535       & 23148       & 17105      & 68166       & 44397       \\
House sales      & King County, US & 2014-2015          & 2098        & 1368       & 6323        & 2770       & 11280       & 3305        \\
Chicago crimes    & Chicago, US & 2022          & 684         & 172        & 3968        & 992        & 20200       & 5051        \\
Philadelphia crimes     & Philadephia, US & 2023    & 425         & 107        & 2363        & 591        & 11512       & 2878        \\
Police Dep. Incidents & San Francisco, US & 2018-2024 & 136         & 35         & 765         & 192        & 3386        & 847         \\ \midrule
Porto taxi   & Porto, Portugal & 2013-2014               & -           & -          &     358958        &          89740  & -           & -           \\
Geolife    & China & 2007-2011                 & -           & -          &      12803       &         3201   & -           &       -     \\ \bottomrule
\end{tabular}
\end{table*}

To construct a comprehensive and representative benchmark, we curate a collection of 7 geospatial datasets that frequently appear in geospatial AI literature and reflect the diversity of real-world spatial data. These datasets encompass various types of geospatial information, including urban infrastructure, human mobility, housing, and public safety. To enhance generalizability and mitigate demographic biases, the selected datasets cover multiple continents, ensuring geographic and socioeconomic diversity across tasks.

\paragraph{Airbnb}: Comprises approximately 3.1 million records from 80 cities worldwide, collected between June 5, 2022, and May 18, 2023. Originating from the Airbnb platform\footnote{ \url{https://insideairbnb.com}}, the dataset includes detailed information about short-term rental listings, including geographic location, property characteristics (e.g., room type, price, availability), host activity, and review metrics. To evaluate spatial modeling in diverse urban contexts, six cities from different countries were selected for inclusion in the benchmark. The dataset was preprocessed to remove incomplete entries and underwent exploratory analysis prior to integration. The dataset is available under the CC BY 4.0 license.

\begin{figure}[t]
    \centering
    \includegraphics[width=0.8\linewidth]{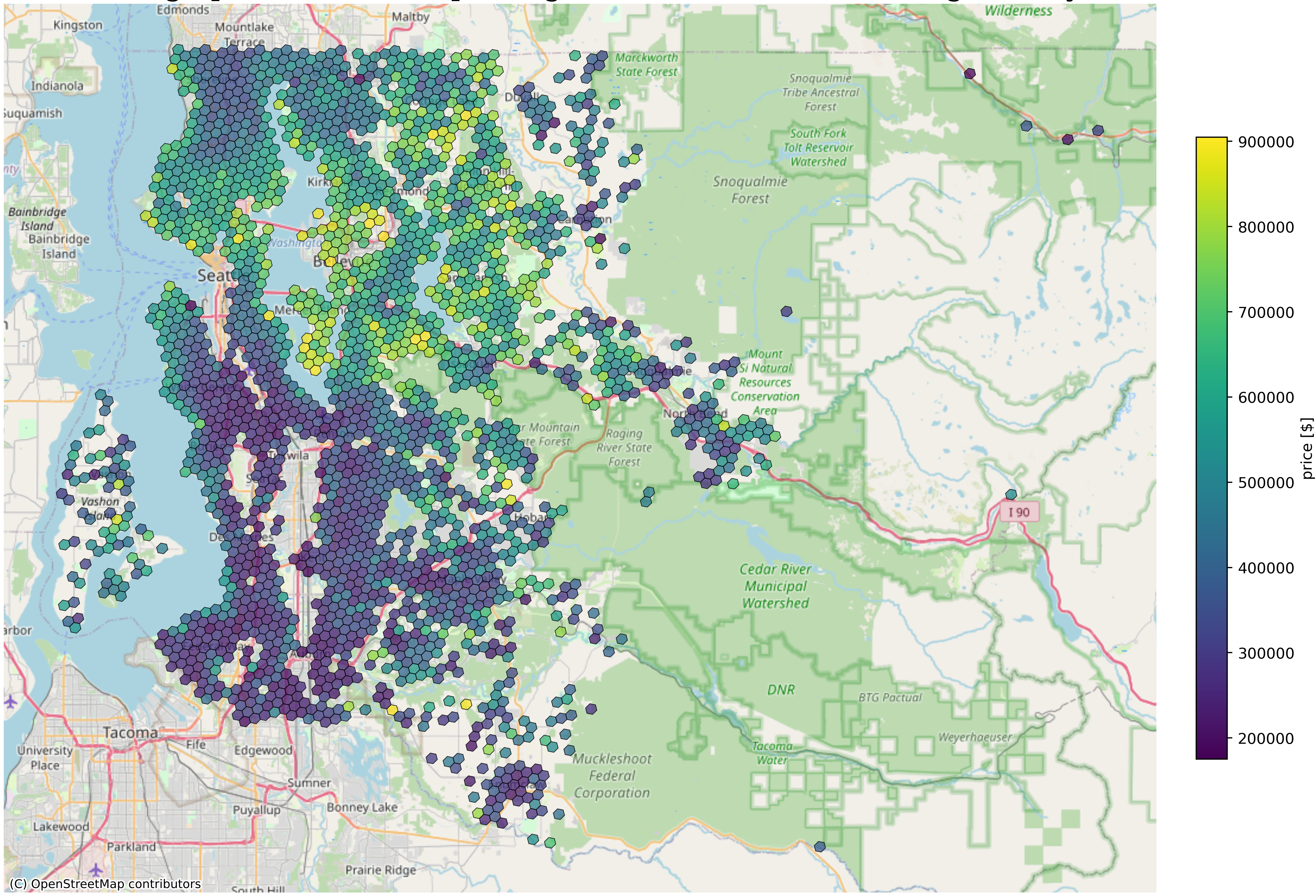}
    \caption{Mean prices per haxagon of houses in King County.}
    \label{fig:house_prices_kings_county}
\end{figure}

\paragraph{House Sales in King County}\cite{housesalesinkingskounty}: Contains approximately 21,000 residential property sales from King County, Washington, recorded between 2014 and 2015. Designed to support the prediction of 2016 housing prices, the dataset includes geographic coordinates, physical property attributes (e.g., size, number of rooms, condition), and contextual features such as proximity to waterfronts. It enables evaluation of spatial information in price prediction tasks, with clear spatial trends—such as higher prices near the bay—demonstrating the significance of location (Figure \ref{fig:house_prices_kings_county}). The dataset is publicly available under the CC0 license.

\paragraph{Chicago Crime Dataset}\cite{chickagocrime}: Comprises over 500,000 reported crime incidents (excluding murders) recorded in the City of Chicago between January 2020 and December 2022. The data originates from the Chicago Police Department’s CLEAR (Citizen Law Enforcement Analysis and Reporting) system and is publicly available under the CC0 license. Prior to inclusion in the benchmark, records falling outside the city limits were removed as outliers. Each entry includes precise time and location details, supporting fine-grained spatial analysis. Theft is the most frequently reported crime, followed by battery, suggesting patterns linked to tourist activity. Seasonal trends show higher crime rates during summer months, highlighting the importance of spatiotemporal modeling in urban safety analysis. For benchmarking purposes, we used a subset of data originating from the year 2022. Additionally, we provide raw data from multiple years without predefined train-test splits.

\paragraph{Philadelphia Crimes Dataset}\footnote{\url{https://opendataphilly.org/datasets/crime-incidents/}}: Includes approximately 1.5 million recorded crime incidents from the city of Philadelphia. Each entry includes the date, time, and type of crime, enabling spatial and temporal analysis. The dataset serves as a complementary resource to the Chicago Crime Dataset, providing an additional urban context for evaluating geospatial models. It is publicly available under City of Philadelphia License. The benchmark relies on a subset of data originating from crime reports in 2023. Due to licensing restrictions, we provide the train-test splits as lists of IDs.
\paragraph{Police Department Incident Reports Dataset}\cite{sf_pdir_2018_2024}: Encompasses over 600,000 reported crime incidents submitted by officers or the public through the city’s online reporting system. Covering the period from January 2018 to March 2024, it is the most recent dataset among those included in the benchmark. Published under the PDDL license. As with the Chicago and Philadelphia datasets, it provides detailed temporal and categorical information on crimes.  

\paragraph{Porto Taxi}\cite{porto_taxi}: Captures detailed GPS trajectory data from 442 taxis operating in Porto, Portugal, over a one-year period between July 1, 2013, and June 30, 2014. The dataset is released under CC BY 4.0 license. It was originally collected to support improvements in electronic taxi dispatch systems. Each record corresponds to a completed taxi trip and includes metadata such as trip origin type, taxi and call identifiers, day type, and a GPS trajectory encoded as a sequence of geographic coordinates. The subset used in the benchmark contains nearly 400,000 individual trajectories, providing a rich resource for modeling spatial and temporal mobility patterns in urban environments.
\paragraph{Geolife}\cite{zheng2009mining, zheng2008understanding, zheng2010geolife}:
Collected by Microsoft Research Asia from 182 users between 2007 and 2012. It contains over 17,000 trajectories and 25 million GPS points. We filter the data to include only points recorded in Beijing. Each point includes geosparial information, timestamp, and transportation mode (e.g., walking, driving, cycling), enabling multimodal mobility analysis. Due to the dataset’s licensing restrictions, we release train-test splits as lists of trajectory IDs, allowing to match them with the original data.

\subsection{Downstream Tasks}
Building upon previously defined objectives and an extensive review of the relevant literature, we introduce 5 downstream tasks designed to reflect practical and theoretically meaningful challenges in the field. These tasks fall into two major categories: (1) region-based prediction tasks, where the objective is to estimate values or labels associated with discrete spatial units such as hexagonal grid cells, and (2) trajectory-based tasks, which involve modeling, forecasting, or classifying dynamic movement patterns over space and time. Together, these categories encompass a diverse range of geospatial problems and provide a comprehensive benchmark for evaluating learned spatial representations, while also offering a solid foundation for future extension and methodological development in geospatial machine learning.
\subsubsection{Region-based tasks}
\paragraph{Short-Term Rental Price Prediction} (\textsc{STRPP}) Short-term rentals have emerged as a significant economic factor in major urban areas in recent years \cite{GARCIALOPEZ2020, celata2022overtourism, semi2021marginal, valentin2021regulating}. This growth has spurred research aimed at understanding and potentially leveraging this sector.  In particular, predicting rental prices in urban areas has become a critical downstream task in real estate and urban planning research \cite{iliopoulou2024spatial, li2019implementation}. To evaluate this task, we propose utilizing data from Airbnb, the widely used short-term rental platform, which offers open access and has been employed in numerous recent studies.

\paragraph{Housing Prices Prediction} (\textsc{HPP}) The prediction of housing prices holds significant societal and economic implications, garnering considerable attention across diverse research disciplines  \cite{glaeser2005have, fuller2020housing}. From a geospatial machine learning perspective, housing prices exhibit strong correlations with a multitude of factors, including location, area, and population demographics, necessitating a comprehensive approach to model individual housing price fluctuations. This complexity has spurred the exploration of advanced machine learning techniques for improved predictive accuracy \cite{mohd2020overview, viana2021attention}. We use the House Sales in King County Dataset to facilitate evaluation and comparative analysis in this domain.

\paragraph{Crime Activity Prediction} (\textsc{CAP}) The increasing availability of extensive crime data has spurred scientific interest in the field of crime analysis and prediction, offering potential to aid law enforcement agencies in proactive crime prevention strategies \cite{buczak2010fuzzy}. Real-world crime patterns exhibit strong correlations with temporal, spatial, and demographic factors, introducing complexity into predictive modeling tasks. Machine learning algorithms, capable of incorporating diverse features such as spatial coordinates, time stamps, and population demographics, have shown promise in enhancing predictive accuracy \cite{bappee2018predicting, xiong2019predicting, suguna2022comparative}. 
To support standardized evaluation and benchmarking in this research domain, we adopt three datasets: the Chicago Crime Dataset, the Police Department Incident Reports Dataset and Philadelphia Crimes Dataset.In our formulation, the predictive task is defined as crime intensity estimation. This can be intuitively interpreted as a measure of relative danger or risk level associated with a given geographic region e.g. neighbourhood.

\subsubsection{Trajectory-based tasks}
\paragraph{Human Mobility Prediction} (\textsc{HMP}) Human mobility prediction, the task of learning and forecasting individuals' movement trajectories within urban environments, is a fundamental component of numerous location-based services and applications \cite{xue2021mobtcast}. This field encompasses the characterization of diverse mobility patterns, such as walking, driving, or utilizing public transportation.  Understanding these patterns is essential for a wide range of domains, including epidemic control \cite{ni2009impact, belik2011natural}, urban planning \cite{zheng2011urban, yuan2012discovering, xin2022vision} and traffic forecasting systems \cite{lana2018road, mathew2021comparative}. In this study, we address the task of human mobility prediction as a autoregressive classification task utilizing widely adopted Geolife and Porto Taxi.

\begin{figure}[t]
    \centering
    \includegraphics[width=0.7\linewidth]{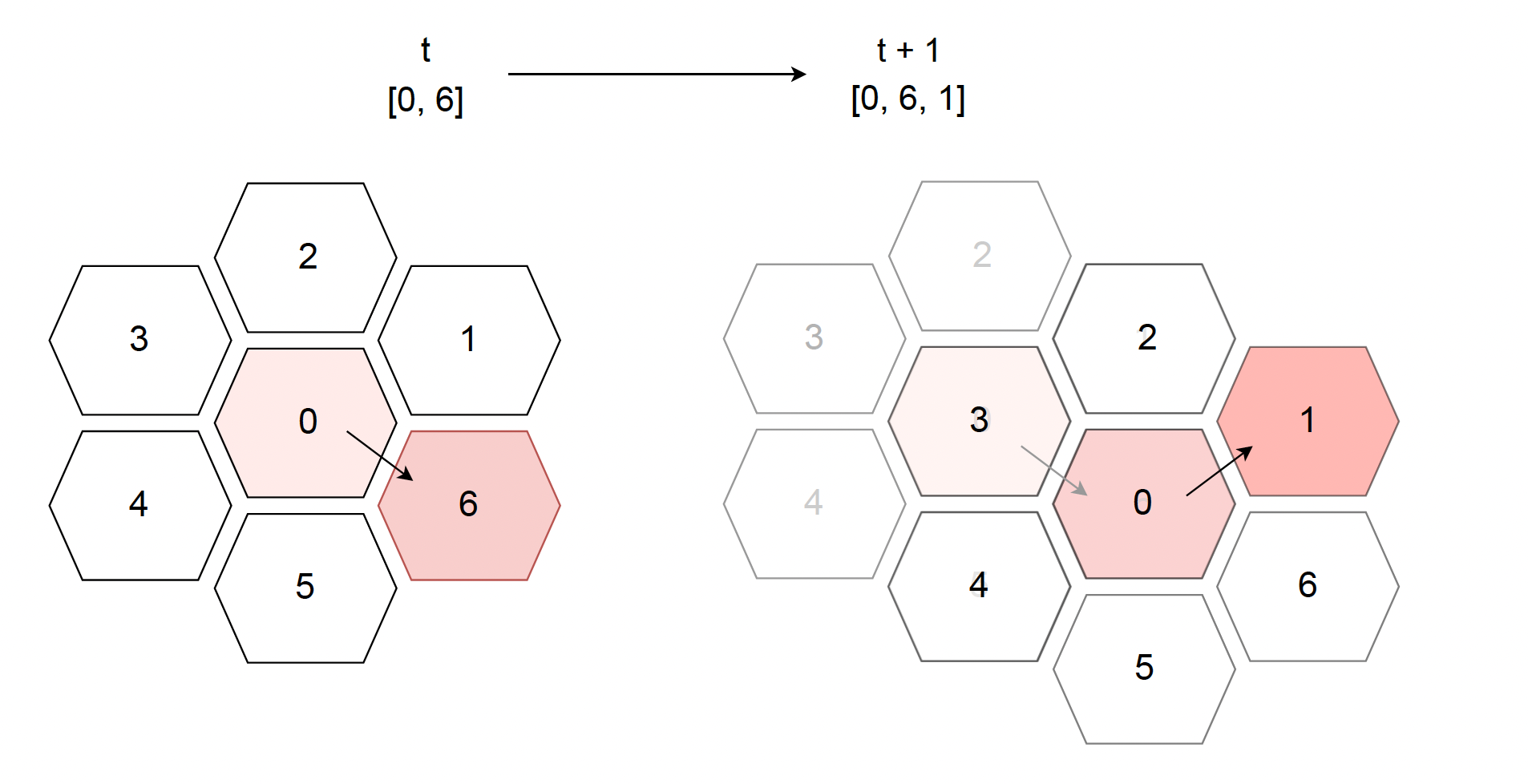}
    \caption{ Spatially consistent class-encoding scheme in which each hexagon is encoded relative to its predecessor.}
    \label{fig:hex}
\end{figure}

\paragraph{Travel Time Estimation} (\textsc{TTE}) is a critical component of modern transportation systems, increasingly utilized as a fundamental service in navigation and intelligent transportation systems \cite{huang2021complementary, jin2021hierarchical}. We propose the utilization of the aformentioned Porto Taxi dataset for time travel estimation analysis as well as Geolife Dataset.

\subsection{Data preprocessing}

Extensive preprocessing was undertaken to ensure data quality, model robustness, and consistency across tasks. This included the removal of outliers, normalization of numerical variables, and the encoding of categorical attributes to ensure standardized input formats suitable for machine learning models. Preprocessing procedures were further tailored to the nature of the task categories. For region-based tasks such as \textsc{STRPP}, \textsc{HPP}, and \textsc{CAP}, spatial point data were aggregated within predefined hexagonal cells. In \textsc{STRPP} and \textsc{HPP}, average target values (pricing) were computed per cell to form the predictive targets. In the case of \textsc{Crime Activity Prediction}, the aggregated count of reported crime incidents within each hexagonal cell was interpreted as a measure of crime intensity, serving as a quantitative proxy for the relative safety of a given area. These intensity values were subsequently normalized to a continuous range 
[0,1] to facilitate comparability and model convergence.

In contrast, for trajectory-based tasks, sequential point data were transformed into paths over a hexagonal spatial grid. While this approach has its advantages and limitations, it allows us to leverage datasets with varying sampling frequencies and temporal resolutions, facilitates spatial aggregation, and improves the comparability of trajectories derived from heterogeneous data sources. Gaps in trajectories—caused by missing or noisy location data—were addressed by estimating intermediate hexagons using shortest-path interpolation and imputing missing feature values based on adjacent valid segments. Furthermore, for \textsc{Human Mobility Prediction} task (next region classification), we adopt a spatially consistent class-encoding scheme in which each hexagon is encoded relative to its predecessor. Spatial transitions are represented as class labels derived from a fixed directional mapping (Figure \ref{fig:hex}), ensuring consistency and enabling the model to learn movement dynamics in a localized and coherent frame of reference .

These preprocessing steps established a coherent spatial structure and temporal consistency essential for downstream modeling. To support robust model evaluation, we devised task-specific data partitioning strategies for each category, ensuring reliable and representative training and testing splits. The details of these splitting methodologies are presented in the following section.

\subsection{Multi Resolution Configuration}

\begin{figure}[t]
    \centering
    \includegraphics[width=0.9\linewidth]{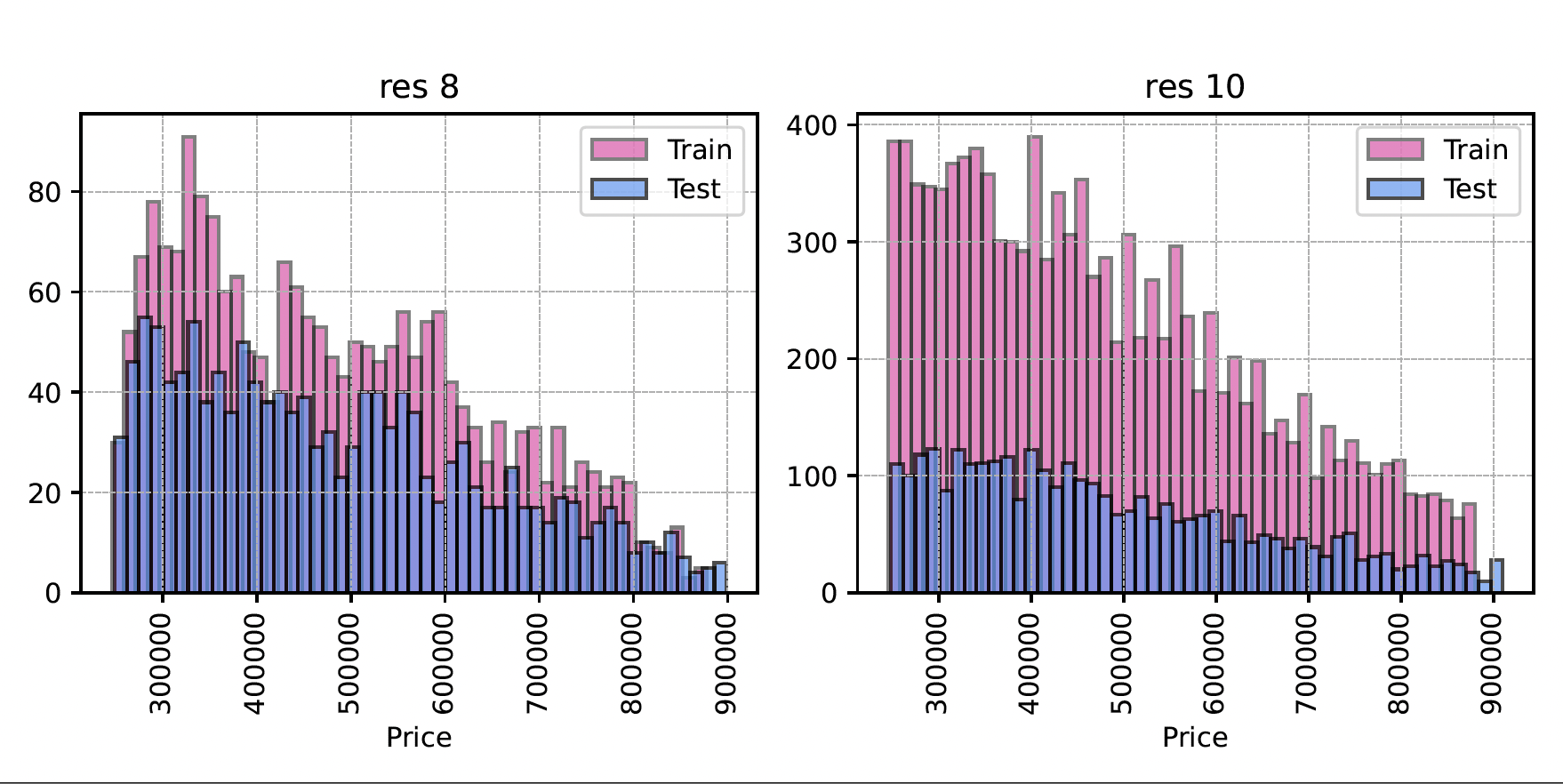}
    \caption{Target feature distribution remains stable for different resolutions in the HPP task.}
    \label{fig:prices_dist}
\end{figure}

\begin{figure}[t]
    \centering
    \includegraphics[width=0.9\linewidth]{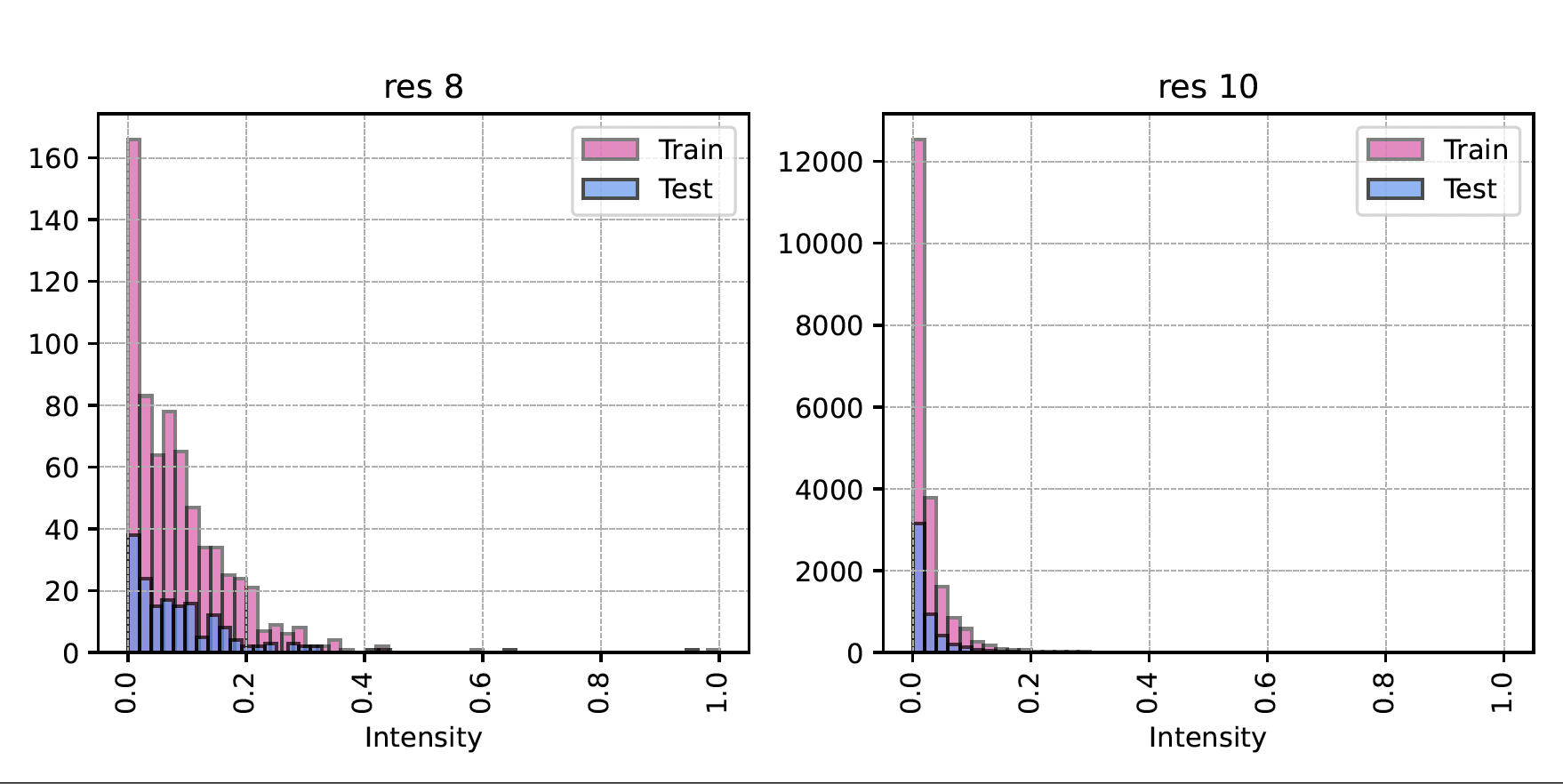}
    \caption{Target feature distribution changes with different resolutions in the CAP task.}
    \label{fig:crimes_dist}
\end{figure}

Building on the motivation for granularity adaptation, the benchmark implements a multi-resolution approach using H3 hierarchical hexagonal grid system to create multiple variants of each geospatial task at different spatial scales. This methodology aggregates data using H3 cells of varying resolutions—from coarse-grained large hexagons to fine-grained small ones—enabling comprehensive evaluation of how well models can handle spatial dependencies across different levels of detail. The impact of resolution changes varies significantly depending on the underlying task characteristics and the nature of the target variable. 
For example, in the \textsc{Housing Prices Prediction} and the \textsc{Short-Term Rental Price Prediction} tasks the target feature distribution typically remains stable across resolutions since the underlying spatial patterns are preserved regardless of aggregation scale (see Figure~\ref{fig:prices_dist}). In contrast, for the task of \textsc{Crime Activity Prediction}, target value systematically decrease as hexagon size decreases, reflecting the mathematical relationship between density measures and the spatial units over which they are calculated (see Figure~\ref{fig:crimes_dist}) This multi-resolution framework ensures that geospatial models are evaluated not just on their ability to process spatial data, but on their robustness to the varying scales of spatial aggregation commonly encountered in real-world applications.

\subsection{Data Splitting}

\begin{figure}[t]
    \centering
    \includegraphics[width=0.9\linewidth, trim=0 0 0 25, clip]{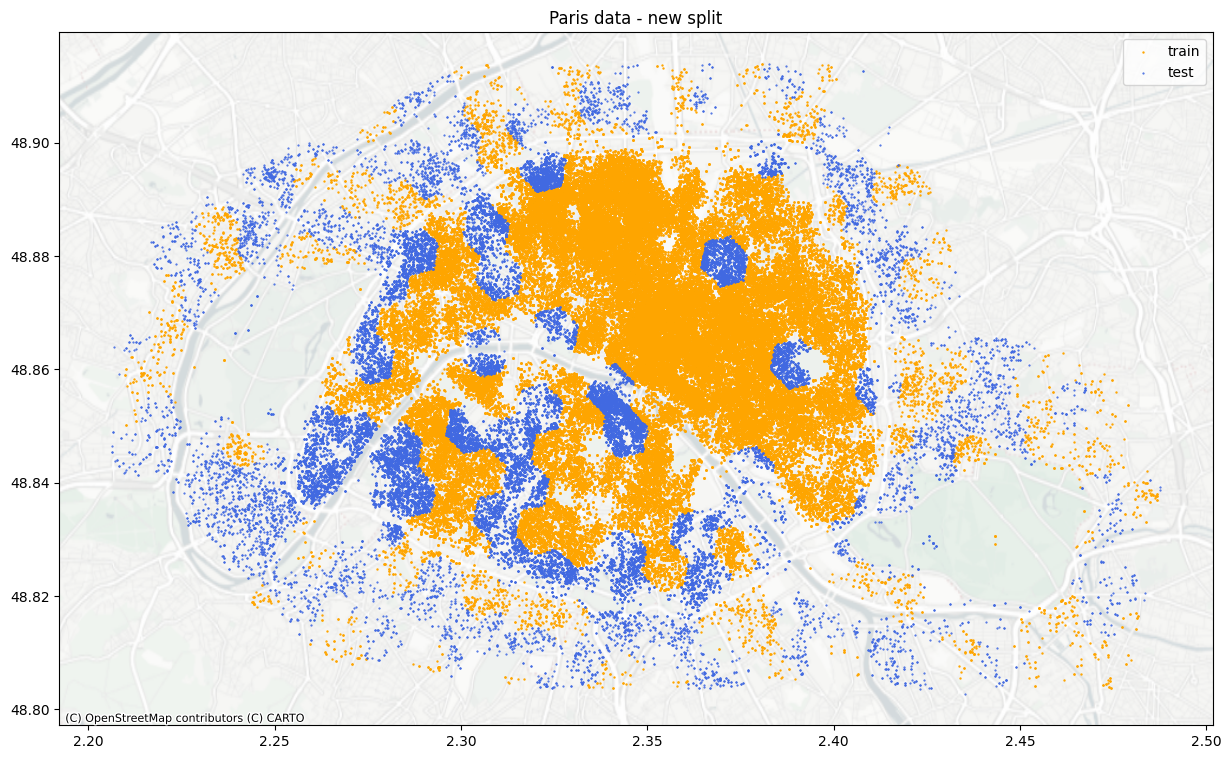}
    \caption{Splitting algorithm ensures no spatial overlap between train and test sets.}
    \label{fig:spatial_split}
\end{figure}

We support a unified strategy for performing train-test splits in geospatial tasks, where entire hexagonal regions—derived from a predefined spatial resolution—are consistently assigned either to the training or testing set. This ensures that no spatial region appears in both splits, preserving the independence of evaluation. Stratification is applied to maintain balanced distributions of the target variable across the splits. When no explicit target variable is provided (e.g. price), we automatically use the number of points within each hexagonal cell as a proxy, treating it as a measure of local data intensity (e.g. in \textsc{CAP} task). Quantile-based bucketing is then applied to achieve balanced splits while respecting spatial boundaries. This method maintains spatial pattern consistency while enabling better local generalization and is used across all tasks to ensure consistent and reproducible evaluations (see Listing~\ref{lst:spatial_points_algorithm} and Figure \ref{fig:spatial_split})

For trajectory-based tasks, splitting is performed based on entire sequences to preserve temporal coherence. In the \textsc{TTE} task, trajectories are stratified by their total travel duration to ensure a balanced range of time spans across the training and testing sets. 
In the \textsc{HMP} task, stratification is based on trajectory length, measured as the number of spatial steps. Each trajectory is subsequently segmented into an input sequence \textit{X} and a target sequence \textit{Y}, where \textit{Y} always corresponds to 15\% of the total trajectory length. The train-test split is stratified by trajectory length to ensure a balanced representation of both short and long movement patterns.

To support reproducibility, we provide a collection of precomputed splits at multiple hexagonal resolutions, ready-to-use splitting methods within our library, and access to raw datasets for customized splitting workflows. Detailed statistics of the proposed splits representing the number of regions and trajectories within the train and test splits, are presented in Table~\ref{tab:splits}. The table also summarizes the corresponding spatial and temporal coverage of each split.

\begin{table*}[t]
\centering
\caption{Performance of different embedders on STRPP task on Airbnb dataset at various resolutions. (H2V - Hex2Vec, GV - GeoVex, CE - CountEmbedder, CCE - ContextualCountEmbedder)}
\label{tab:airbnb}
\renewcommand{\arraystretch}{0.75}  
\begin{tabular}{l|rrrr|rrrr|rrrr}
\toprule
& \multicolumn{4}{c}{\textbf{Res 8}} 
    & \multicolumn{4}{c}{\textbf{Res 9}} 
    & \multicolumn{4}{c}{\textbf{Res 10}} \\
\textbf{Metric} & \multicolumn{1}{c}{H2V} & \multicolumn{1}{c}{GV} & \multicolumn{1}{c}{CCE} & \multicolumn{1}{c}{CE}
    & \multicolumn{1}{c}{H2V} & \multicolumn{1}{c}{GV} & \multicolumn{1}{c}{CCE} & \multicolumn{1}{c}{CE}  
    & \multicolumn{1}{c}{H2V} & \multicolumn{1}{c}{GV} & \multicolumn{1}{c}{CCE} & \multicolumn{1}{c}{CE}  \\
\midrule
MSE $\times10^2$
      & 40.10 & 48.33 & 40.21 &  \textbf{38.30}
      & 45.97 & 50.57 &  45.65 & \textbf{42.53}
      & 59.48 & 63.80 & \textbf{44.96} & 66.27\\
RMSE  & 63.33   & 69.53   & 63.41  &  \textbf{61.89}  
      & 67.80   & 71.11   &  67.50 & \textbf{65.22} 
      & \textbf{77.12}   & 79.87   & 81.83  & 81.40 \\
MAE   & 43.03   & 48.58  & \textbf{39.69}  &  \textbf{39.61  } 
      & 46.97   & 50.75   & \textbf{42.98}   & 43.21   
      & \textbf{54.55}   & 56.42   & 56.96   & 57.08 \\
MAPE  & 51.38  & \textbf{47.47}   & 57.26  & 58.16 
      & 54.17   & \textbf{53.90}   &  60.09  & 60.90
      & 53.02   & \textbf{51.76}   &  53.53  & 54.48\\
sMAPE & 44.49  & \textbf{41.39}   & 48.09  &  47.88
      & 45.85   & \textbf{44.67 } & 48.67  &  49.05  
      & \textbf{46.76}  & \textbf{46.76}   &  49.13 & 49.03\\
\bottomrule
\end{tabular}
\end{table*}

\begin{table*}[t]
\centering
\caption{Performance of different embedders on HPP task on House Sales in King County dataset at various resolutions. (H2V - Hex2Vec, GV - GeoVex, CE - CountEmbedder, CCE - ContextualCountEmbedder)}
\label{tab:HouseSales}
\renewcommand{\arraystretch}{0.75}  
\begin{tabular}{l|rrrr|rrrr|rrrr}
\toprule
& \multicolumn{4}{c}{\textbf{Res 8}} 
    & \multicolumn{4}{c}{\textbf{Res 9}} 
    & \multicolumn{4}{c}{\textbf{Res 10}} \\
\textbf{Metric} & \multicolumn{1}{c}{H2V} & \multicolumn{1}{c}{GV} & \multicolumn{1}{c}{CCE} & \multicolumn{1}{c}{CE} 
    & \multicolumn{1}{c}{H2V} & \multicolumn{1}{c}{GV} & \multicolumn{1}{c}{CCE} & \multicolumn{1}{c}{CE}  
    & \multicolumn{1}{c}{H2V} & \multicolumn{1}{c}{GV} & \multicolumn{1}{c}{CCE} & \multicolumn{1}{c}{CE}  \\
\midrule
MSE $\times10^9$  & 27.51 & \textbf{26.98} & 31.69 & 58.06
                   & \textbf{23.18} & 26.82 & 28.16 & 26.83 
                   & 25.28 & 28.19 & \textbf{22.22} & 25.22 \\
RMSE $\times10^4$ & 
16.577& \textbf{16.42} & 17.80 & 24.09 
      & \textbf{15.22} & 16.37 & 16.77 & 16.37 
      & 15.90 & 16.78 &  \textbf{14.90} & 15.83 \\
MAE $\times10^4$  & \textbf{12.62} & 13.41 & 12.95 & 18.10 
      & \textbf{11.90} & 13.38 & 12.53 & 12.69 
      & 12.56 & 13.23 &  \textbf{11.83} & 12.62\\
MAPE  & 33.91 & \textbf{30.42} & 40.26 & 43.35 
      & 33.28 & \textbf{31.64} & 37.51 & 33.75
      & 32.22 & \textbf{31.40} & 34.64 & 32.68 \\
sMAPE & 33.73 & \textbf{29.23} & 38.95 & 46.99 
      & 31.67 & \textbf{29.81} & 36.25 & 32.22
      & \textbf{30.57} & 30.69 & 31.81 & 30.68  \\
\bottomrule
\end{tabular}
\end{table*}

\section{Evaluation}

\subsection{Baseline models}
To provide reproducible reference points for evaluating spatial representations, we implement a set of simple baseline models. These models \textit{are not} designed to achieve state-of-the-art performance, but rather to serve as standardized evaluation tools within the benchmark. The primary objective of the provided baselines is to enable the use of \textit{any} available spatial embeddings and to facilitate direct comparisons between them. However, when the focus shifts from benchmarking embeddings to maximizing task-specific performance, these baseline models can be extended or replaced with more sophisticated architectures or end-to-end solutions, and additional metadata from the datasets may be incorporated to improve results.

For price prediction tasks, we employ a feedforward neural network with multiple linear layers and ReLU activations. Conversely, for \textsc{CAP} tasks, sigmoid activation is utilised, with the objective being to attain a range of [0, 1]. For sequence-based predictions we employ a recurrent architectures based on LSTM layers, which captures temporal dependencies across variable-length sequences of spatial embeddings. In the case of the \textsc{Travel Time Estimation} task, the model performs regression based on the entire input sequence. In contrast, for the \textsc{Human Mobility Prediction} task, the model outputs a class label at each timestep. To enhance sequence modeling in the latter case, we adopt teacher forcing algorithm \cite{lamb2016professorforcingnewalgorithm} during training, wherein ground-truth values are used to guide the model’s predictions at each step. Additionally, we incorporate a multihead attention layer to effectively capture temporal dependencies and contextual relationships across the entire trajectory. Further details regarding the model architecture and hyperparameters are provided in the appendix \ref{appendix:hyper}, while the complete implementation is publicly available through the \textsc{SRAI} library.

\begin{lstlisting}[language=Python, caption=Point-based train-test splitting methods., label=lst:spatial_points_algorithm]
Input: Point dataset G, H3 resolution r, optional target y, number of bins n

# 1. Regionalize space into hexagonal cells
Assign each point p in G to an H3 cell h at resolution r

# 2. Define stratification variable
If y is provided:
    Use y values (categorical or numeric) to guide stratification
Else:
    Use point counts per H3 cell

# 3. Bucket data for balanced splits
If numeric y or counts:
    Divide values into n quantile-based bins

# 4. Split into train and test sets
Stratify data based on buckets to ensure balanced distribution
Assign all points from each H3 cell to the selected split

Return train and test datasets
\end{lstlisting}

\subsection{Evaluation methods}
To ensure comparable assessment of spatial representations and model performance across the benchmark, we define tailored evaluation strategies for both regression-based and trajectory-based tasks.

For regression tasks, we employ standard error-based evaluation metrics, including the Mean Squared Error (MSE), Root Mean Squared Error (RMSE), and Mean Absolute Error (MAE). In addition, to capture relative predictive performance, we include the Mean Absolute Percentage Error (MAPE), which expresses error as a percentage of the true value. To address the limitations of MAPE in cases where target values approach zero, we also compute the Symmetric Mean Absolute Percentage Error (sMAPE), which symmetrically penalizes over- and under-predictions. These metrics are employed for both the \textsc{STRPP} and \textsc{HPP} tasks, as well as for the \textsc{TTE} task, which—despite being trajectory-based—is also formulated as a regression problem. In the context of the \textsc{CAP} task, where the target variable represents a normalized crime intensity value constrained to the [0, 1] range, percentage-based metrics such as MAPE and sMAPE become unstable and potentially misleading due to their sensitivity to small denominators. Consequently, they are excluded in this particular task and instead the Coefficient of Determination, denoted by $R^2$, is incorporated. The $R^2$ metric is a quantitative measure of the proportion of variance in the dependent variable that is predictable from the independent variables. It offers an interpretable measure of goodness-of-fit for bounded regression tasks.

For \textsc{HMP} task, which require accurate prediction of spatial sequences, we introduce a combination of pointwise and sequence-level metrics:
\paragraph{Average Haversine Distance}: Computes the mean pairwise great-circle distance in meters between corresponding predicted and true H3 cell centroids within each trajectory. This offers a spatially meaningful measure of local deviation in the predicted path.
\paragraph{Dynamic Time Warping (DTW) Distance}: Quantifies the temporal and spatial misalignment between predicted and ground-truth trajectories. Originally developed for time series analysis, DTW \cite{DTW} allows flexible, non-linear alignment of sequence elements, capturing global shape similarity. In benchmark, we compute the pairwise distances within DTW using the Haversine formula. It is particularly valuable when evaluating predictions where the same destination is reached via slightly different pacing or intermediate steps, making it a complementary metric to average Haversine distance.
\paragraph{Sequence Accuracy}: Measures the proportion of exact element-wise matches between predicted and ground-truth H3 index sequences. This metric provides a straightforward assessment of how often the model predicts the correct location at each step. While intuitive, it is a strict metric—even small spatial or temporal deviations result in lower scores, despite the overall trajectory being reasonably accurate.

As longer sequences naturally increase prediction difficulty, we perform evaluation with these metrics over the first \textit{k} steps of target trajectory. This approach allows us to granulate task difficulty and better capture model performance at different stages of the prediction horizon.

All of the above, including embedders, task-oriented baseline models and evaluation methods, are fully implemented and readily available as a module in the \texttt{SRAI} library, allowing for consistent, reproducible evaluation across all benchmark tasks. Moreover, they are tightly integrated with the proposed train–test splits for each dataset, enabling researchers to efficiently compute evaluation results of novel geospatial representation learning methods.

\subsection{Results}

In this Section, we present a comparative evaluation of four geospatial embedding models across multiple spatial analysis tasks, utilizing benchmark datasets of varying resolutions. The reported metrics reflect the average performance across multiple experimental runs to ensure robustness and comparability. In this study, we focus on OSM-based embedders due to the current integration with the \textsc{SRAI} library, which provides native support for these models. However, the benchmark framework is designed to be fully extensible, allowing the inclusion and comparison of diverse embedders from any modality. The evaluated embedders include:
\begin{itemize}

\item \textbf{Hex2Vec} (H2V) \cite{Wo_niak_2021}, a Skip-gram model with negative sampling, adopted for region embedding pre-training; 
\item \textbf{CountEmbedder} (CE) \cite{raczycki_ce_cce}, a frequency-based representation model that encodes OSM features; 
\item \textbf{ContextualCountEmbedder} (CCE) \cite{raczycki_ce_cce}, an extension of CountEmbedder that incorporates local spatial context through neighborhood aggregation; 
\item \textbf{GeoVex} (GV) \cite{geovex}, a convolution-based geospatial auto-encoder, which learns region representations from OSM features. 

\end{itemize}
All of the embedding models work on Uber's H3 cells as regions. Each embedder was applied using consistent preprocessing settings, and with specific parameters that are referenced in Appendix~\ref{appendix:hyper}. Additional dataset features, such as categorical and numerical variables, were deliberately excluded to enable an isolated assessment of geospatial embedding performance, uncontaminated by the influence of other data modalities.

One of the first insights that emerges from the region-based tasks is the critical role of multi-resolution settings. Certain embedders achieve higher accuracy at fine-grained resolutions, yet their performance declines as the spatial units become larger (Table \ref{tab:airbnb}, \ref{tab:CAP}), revealing that scalability and spatial sensitivity remain open challenges in geospatial representation learning. This observation underscores the challenges associated with scalability and spatial sensitivity in the context of geospatial learning. As the spatial units increase in size, they have a tendency to aggregate heterogeneous signals. While a proportion of this information may prove useful, a significant amount can become irrelevant or noisy, depending on the nature of the task at hand. In contrast, finer resolutions may necessitate models to capture subtle local variations. Therefore, the ability of an embedder to selectively encode relevant features becomes increasingly important as resolution changes. The current embedding structures and assumptions are sufficiently straightforward in nature to provide a valuable foundation; however, the findings underscore the necessity for more sophisticated, task-specific spatial representation learning methods that can adapt more effectively across diverse scales.

\begin{figure}[t]
    \centering
    \begin{minipage}{0.44\columnwidth}
        \centering
        \includegraphics[width=\textwidth, trim=0 0 0 50, clip]{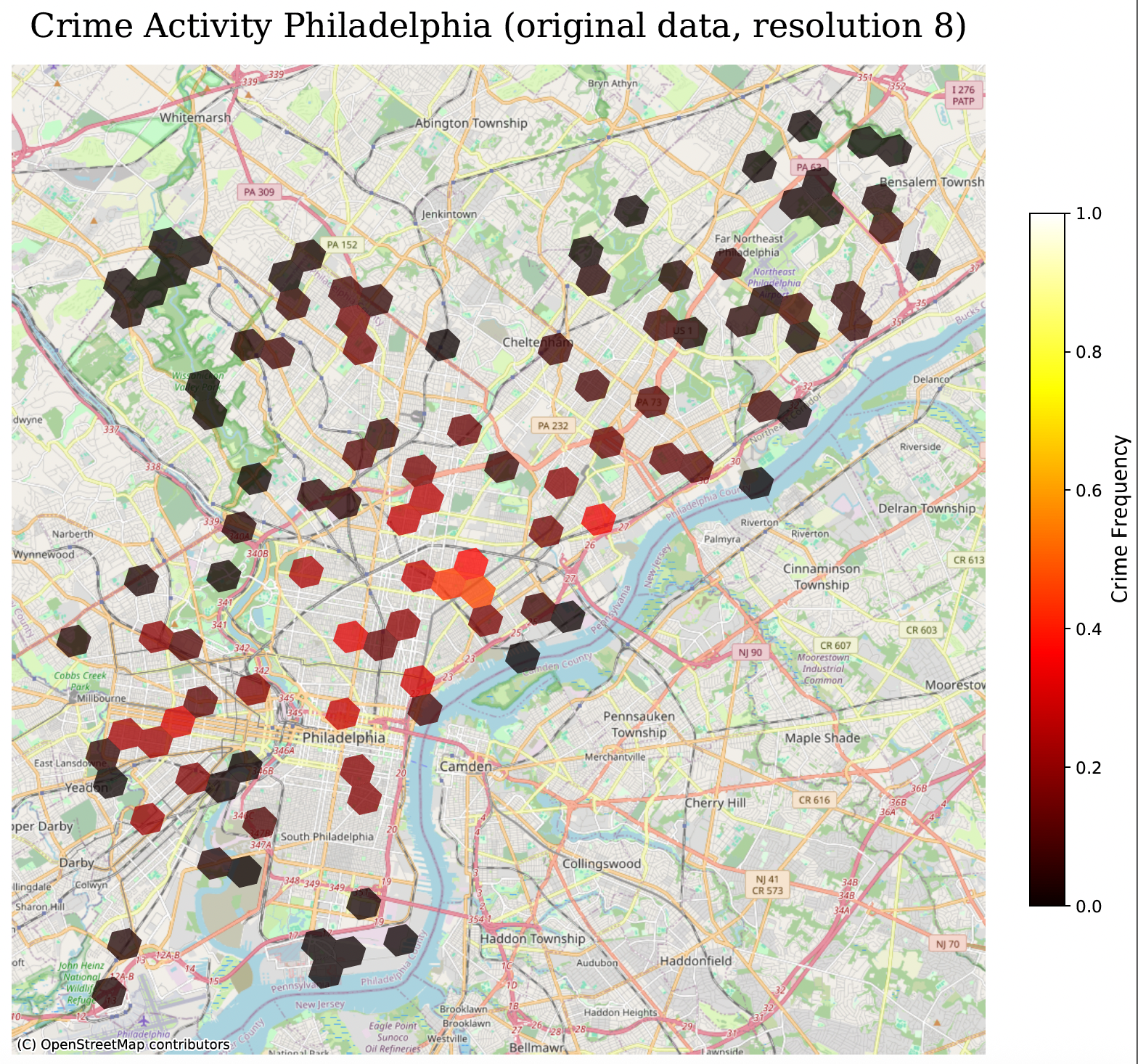}
        \caption*{(a) Original}
    \end{minipage}%
    \hfill
    \begin{minipage}{0.44\columnwidth}
        \centering
        \includegraphics[width=\textwidth, trim=0 0 0 50, clip]{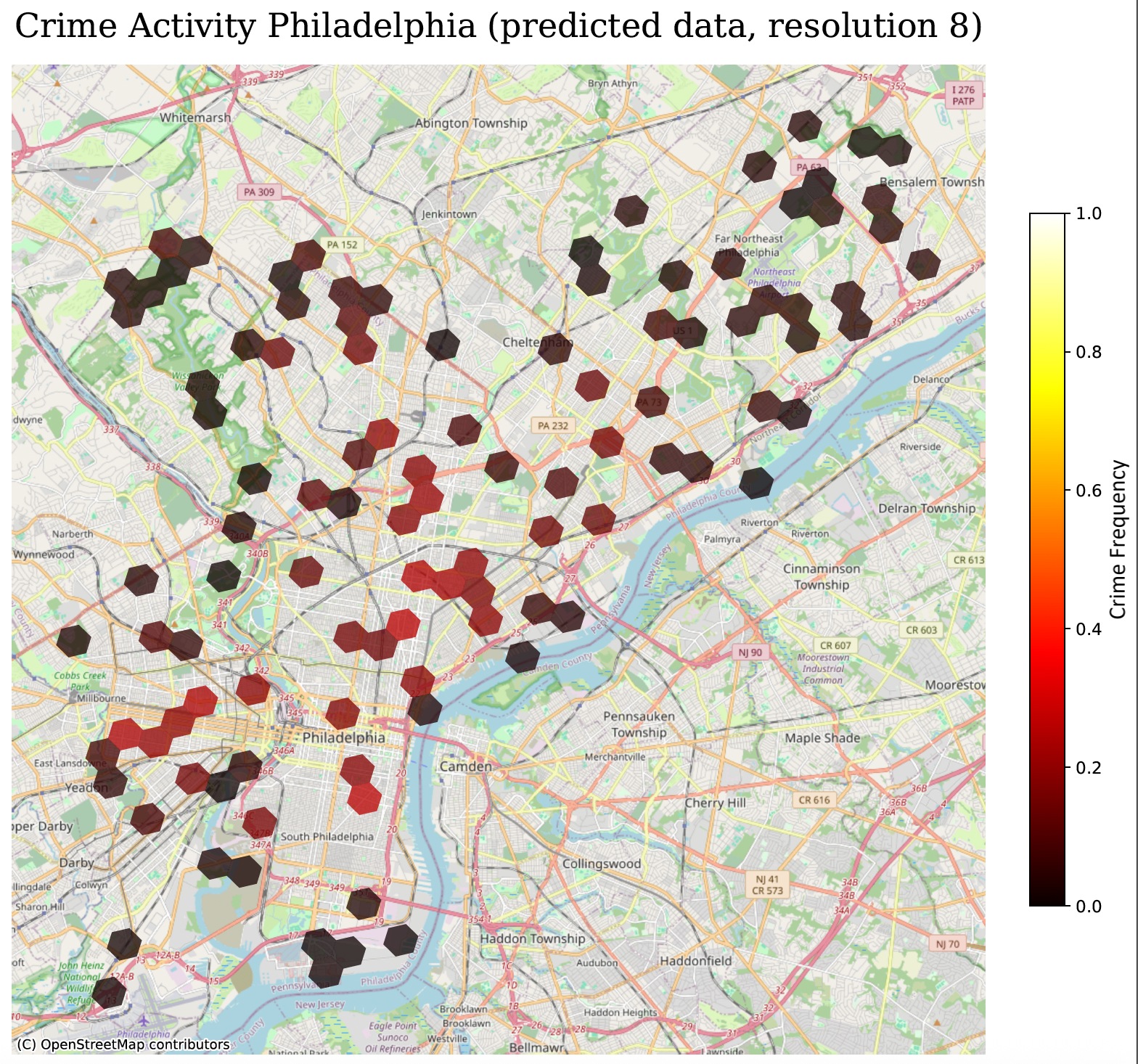}
        \caption*{(b) Predictions}
    \end{minipage}

    \caption{Crime Activity Prediction task on Philadelphia Crimes Dataset using Hex2Vec embeddings.}
    \label{fig:crimes_prediction}
\end{figure}

It can be observed that, in the case of the \textsc{Criminal Activity Prediction} task being treated as an intensity prediction within a normalized 0–1 range, the absolute differences between model outputs are naturally constrained, potentially diminishing the significance of performance comparisons. As illustrated in Fig \ref{fig:crimes_dist}, the distribution of target values is skewed, with higher-intensity crime regions being significantly underrepresented in the dataset. The issue becomes more pronounced at coarser spatial resolutions. This imbalance introduces an additional challenge for learning algorithms, which may bias toward predicting low to mid-range intensities to minimize global error (Fig \ref{fig:crimes_prediction}, Table \ref{tab:CAP}). Thus, we highlight the importance of using relative improvements and statistical validation to interpret the results meaningfully.

\begin{figure}[t]
    \centering
    \begin{minipage}{0.49\columnwidth}
        \centering
        \includegraphics[width=\textwidth, trim=0 2 25 0, clip]{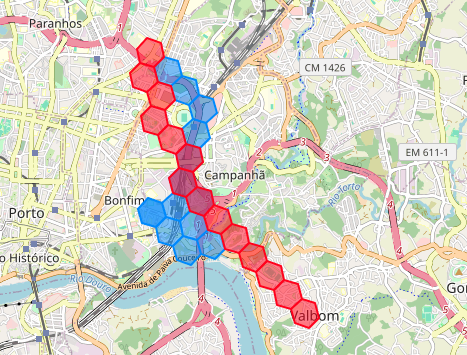}
        \caption*{(a) GeoVex, Porto Taxi Dataset}
    \end{minipage}%
    \hfill
    \begin{minipage}{0.49\columnwidth}
        \centering
        \includegraphics[width=\textwidth, trim=0 2 0 30, clip]{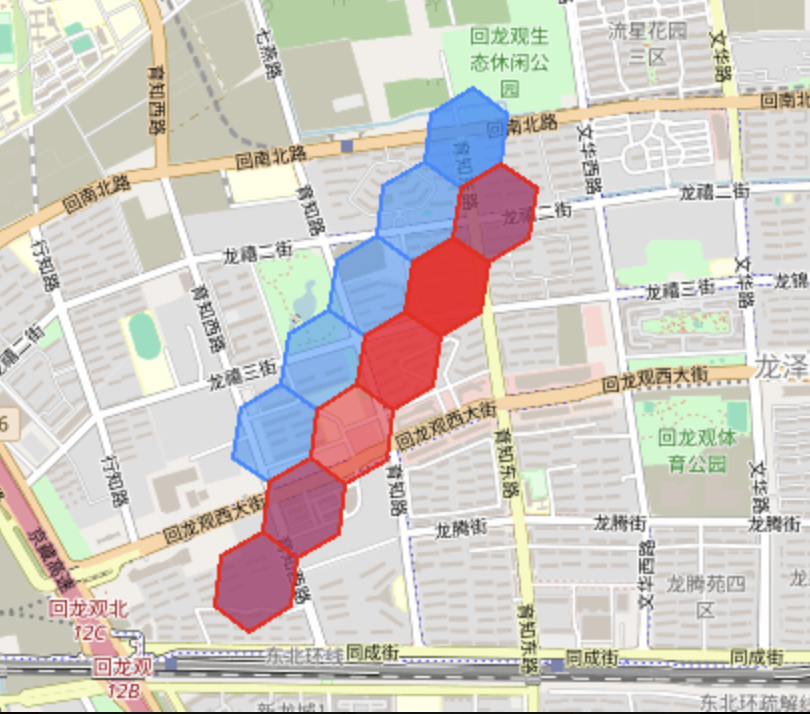}
        \caption*{(b) Hex2Vec, Geolife Dataset}
    \end{minipage}

    \caption{Human Mobility Prediction task. Predicted sequences in red, gold-answers in blue.}
    \label{fig:traj-example}
\end{figure}

It is also crucial to underline that embedding models evaluated on this benchmark operate with deliberately constrained input data—utilizing only OSM tag counts from regions, with some models incorporating neighborhood context. While these models lack additional modalities such as satellite imagery or street-view data, they serve as essential baselines for establishing the performance floor that more sophisticated multi-modal approaches should exceed. This baseline approach provides valuable insights into how much performance gain can be attributed specifically to multi-modality versus spatial reasoning capabilities alone. OSM-based feature counts represent a particularly suitable baseline because they offer structured, globally available geospatial information that other modalities typically enhance rather than replace, making performance comparisons meaningful and interpretable. The limitations of these count-based embeddings become most apparent in trajectory-related tasks (see Table \ref{tab:tte} and Table \ref{tab:hmp}), where the combination of static regional features with basic sequence models like LSTMs fails to capture the dynamic spatial relationships and movement semantics that define realistic trajectories. This shortcoming is exemplified in Figure~\ref{fig:traj-example}, where the original trajectory clearly follows established highway infrastructure, but the model's predictions deviate significantly from road networks, ultimately crossing through forested areas where no viable transportation routes exist.

\begin{table}[ht]
\centering
\caption{Performance of different embedders on the TTE task for Geolife and Porto Taxi datasets (resolution 9). H2V - Hex2Vec, GV - GeoVex, CE - CountEmbedder, CCE - ContextualCountEmbedder.}
\label{tab:tte}
\renewcommand{\arraystretch}{0.7}  
\begin{tabular}{l|rrrr}
\toprule
\multicolumn{5}{c}{\textbf{Geolife}} \\
\textbf{Metric} & \multicolumn{1}{c}{H2V} & \multicolumn{1}{c}{GV} & \multicolumn{1}{c}{CCE} & \multicolumn{1}{c}{CE} \\
\midrule
MSE  $\times10^{5}$ & 2324.434 & 2336.848 & 2336.848 & 2334.396 \\
RMSE $\times10^{3}$ & 15.246   & 15.286   & 15.286   & 15.278 \\
MAE  $\times10^{2}$ & 76.201   & 76.653   & 76.653   & 76.549 \\
MAPE                & 71.194   & 69.676   & 69.676   & 68.356 \\
\midrule
\multicolumn{5}{c}{\textbf{Porto Taxi}} \\

\midrule
MSE  $\times10^{5}$ & 1.037    & 0.954    & 0.881    & 0.850 \\
RMSE $\times10^{3}$ & 0.320    & 0.308    & 0.296    & 0.291 \\
MAE  $\times10^{2}$ & 1.932    & 1.911    & 1.782    & 1.755 \\
MAPE                & 37.233   & 38.863   & 39.550   & 40.882 \\
\bottomrule
\end{tabular}
\end{table}

\begin{table*}[ht]
\centering
\caption{Performance of different embedders on CAP task on Chicago Crimes (CC), Philadelphia Crimes (PC) and Police Department Incidents (PDI) datasets at various resolutions. H2V - Hex2Vec, GV - GeoVex, CE - CountEmbedder, CCE - ContextualCountEmbedder}
\label{tab:CAP}
\renewcommand{\arraystretch}{1}  
\begin{tabular}{l|rrrr|rrrr|rrrr|c}
\toprule
& \multicolumn{4}{c|}{\textbf{Res 8}} 
  & \multicolumn{4}{c|}{\textbf{Res 9}} 
  & \multicolumn{4}{c|}{\textbf{Res 10}} 
  &  \\
\textbf{Metric} 
  & \multicolumn{1}{c}{H2V} & \multicolumn{1}{c}{GV} & \multicolumn{1}{c}{CE} & \multicolumn{1}{c}{CCE} 
  & \multicolumn{1}{c}{H2V} & \multicolumn{1}{c}{GV} & \multicolumn{1}{c}{CE} & \multicolumn{1}{c}{CCE}
  & \multicolumn{1}{c}{H2V} & \multicolumn{1}{c}{GV} & \multicolumn{1}{c}{CE} & \multicolumn{1}{c}{CCE}
  & \\

\midrule
MSE  $\times10^{-3}$  & 10.093 & 13.787 & 7.694 & \textbf{5.840} & 2.959 & 4.063 & 2.475 & \textbf{1.965} & 1.922 & 1.984 & \textbf{1.836} & \textbf{1.778} & \multirow{4}{*}{\rotatebox{90}{\textbf{CC}}} \\
RMSE  $\times10^{-2}$ & 10.046 & 11.741 & 8.759 & \textbf{7.635} & 5.440 & 6.374 & 4.974 & \textbf{4.431} & 4.384 & 4.454 & \textbf{4.285} & \textbf{4.216} & \\
MAE  $\times10^{-3}$   & 50.685 & 59.865 & 44.348 & \textbf{37.100} & 30.952 & 36.834 & 29.442 & \textbf{24.698} & 20.011 & 20.629 & 20.531 & \textbf{19.935} & \\
R2 $\uparrow$   & 0.223 & -0.060 & 0.408 & \textbf{0.550} & 0.339 & 0.093 & 0.447 & \textbf{0.561} & 0.080 & 0.050 & 0.121 & \textbf{0.149} & \\
\midrule

MSE  $\times10^{-3}$  & 5.458 & 4.445 & 3.869 & \textbf{3.422} & 4.327 & 4.762 & \textbf{3.528} & \textbf{3.507} & 2.104 & 2.162 & 0.827 & \textbf{0.074} & \multirow{4}{*}{\rotatebox{90}{\textbf{PC}}} \\
RMSE $\times10^{-2}$  & 7.372 & 6.675 & 6.214 & \textbf{5.848} & 6.578 & 6.901 & \textbf{5.937} & \textbf{5.922} & 4.587 & 4.650 & 2.876& \textbf{2.723} & \\
MAE  $\times10^{-3}$   & 46.967 & 43.667 & 39.095 & \textbf{37.989} & 30.983 & 34.039 & 29.811 & \textbf{27.892} & 18.569 & 18.888 & \textbf{11.549} & \textbf{11.506} & \\
R2 $\uparrow$    & 0.498 & 0.590 & \textbf{0.644} & \textbf{0.685} & 0.325 & 0.257 & \textbf{0.449} & \textbf{0.453} & 0.075 & 0.049& 0.098 & \textbf{0.191} & \\
\midrule
MSE  $\times10^{-3}$  & 20.691 & 21.111 & 20.591 & \textbf{19.024} & 6.544 & 5.978 & 4.427 & \textbf{3.543} & 4.591 & 4.899 & \textbf{3.420} & 3.724 & \multirow{4}{*}{\rotatebox{90}{\textbf{PDI}}} \\
RMSE  $\times10^{-2}$ & 14.384 & 14.533 & 14.348 & \textbf{13.784} & 8.089 & 7.733 & 6.652 & \textbf{5.950} & 6.775 & 6.999 & \textbf{5.848} & 6.102 & \\
MAE  $\times10^{-3}$   & 70.162 & 70.633 & 69.485 & \textbf{64.287} & 31.947 & 31.689 & 28.400 & \textbf{24.798} & 27.626 & 27.411 & 24.297 & \textbf{23.951} & \\
R2 $\uparrow$  & -0.105 & -0.128 & -0.100 & \textbf{-0.016} & 0.335 & 0.393 & 0.550 & \textbf{0.640} & 0.204 & 0.151 & 0.287 & \textbf{0.354} & \\
\bottomrule
\end{tabular}
\end{table*}

\begin{table*}[ht]
\centering
\caption{Performance of different embedders on HMP task for Geolife and Porto Taxi datasets (resolution 9) for different length of sequence \textit{k}. H2V - Hex2Vec, GV - GeoVex, CE - CountEmbedder, CCE - ContextualCountEmbedder}
\label{tab:hmp}
\renewcommand{\arraystretch}{1}
\begin{tabular}{l|rrrr|rrrr|c}
\toprule
& \multicolumn{4}{c|}{\textbf{Geolife}} 
  & \multicolumn{4}{c|}{\textbf{Porto Taxi}} 
  &  \\
\textbf{Metric (avg)} 
  & \multicolumn{1}{c}{H2V} & \multicolumn{1}{c}{GV} & \multicolumn{1}{c}{CE} & \multicolumn{1}{c}{CCE} 
  & \multicolumn{1}{c}{H2V} & \multicolumn{1}{c}{GV} & \multicolumn{1}{c}{CE} & \multicolumn{1}{c}{CCE}
  & \textbf{$k$} \\
\midrule
H Dist. [m]     & 345.44 & 354.19 & 309.77 & 311.67 
                & 397.99 & 439.71 & 400.35 & 404.68 
                & \multirow{3}{*}{1}\\
DTW Dist. [m]   & 345.44 & 354.19 & 309.77 & 311.67
                & 397.99 & 439.71 & 400.35 & 404.68 
                &  \\
Seq Acc [\%] 
$\uparrow$      & \textbf{15.203} & 11.843 & 8.849 & 8.383 
                & 9.267 & \textbf{13.865} & 10.507 & 9.785 
                & \\
\midrule

H Dist. (m)     & 554.02 & 579.21 & 449.84 & 452.79 
                & 571.11  & 644.46  & 573.23  & 576.17  
                & \multirow{3}{*}{3}\\
DTW Dist. [m]   & 1544.26 & 1614.76 & 1247.07 & 1252.05 
                & 1530.98 & 1955.50 & 1542.14 & 1550.60 
                &  \\
Seq Acc  [\%] 
$\uparrow$      & \textbf{7.640} & 6.637 & 6.531 & 5.855 
                & 6.584 & \textbf{7.488} & 7.228 & 7.111 
                & \\
\midrule
H Dist. [m]     & 705.53 & 753.66 & 564.72 & 565.68 
                & 630.75  & 715.66  & 633.78  & 635.68  
                & \multirow{3}{*}{5}\\
DTW Dist. [m]   & 3049.01 & 3262.59 & 2430.33 & 2422.46 
                & 2068.61 & 2716.54 & 2084.27 & 2094.27 
                &  \\
Seq Acc [\%] 
$\uparrow$      & \textbf{5.806} & 5.088 & 5.562 & 4.879 
                & 6.243 & \textbf{6.949} & \textbf{6.825} & 6.715 
                & \\
\midrule
H Dist. [m]     & 831.09 & 898.90 & 658.36 & 658.47 
                & 654.19  & 745.39  & 656.64  & 658.50  
                & \multirow{3}{*}{7}\\
DTW Dist. [m]   & 4825.34 & 5230.19 & 3807.40 & 3793.70 
                & 2245.80 & 2958.89 & 2258.67 & 2268.14 
                &  \\
Seq Acc  [\%]
$\uparrow$      & \textbf{5.070} & 4.537 & \textbf{5.134} & 4.477 
                & 6.175 & \textbf{6.866} & \textbf{6.746} & 6.640 
                & \\
\midrule
H Dist. [m]     & 976.77 & 1063.70 & 762.80 & 761.85 
                & 665.10  & 761.06  & 667.79  & 669.39  
                & \multirow{3}{*}{10} \\
DTW Dist. [m]   & 7693.66 & 8316.19 & 5970.37 & 5941.68 
                & 2252.10 & 2967.73 & 2264.44 & 2274.48 
                &  \\
Seq Acc [\%] 
$\uparrow$      & 4.545 & 4.210 & \textbf{4.844} & 4.243 
                & 6.159 & \textbf{6.842} & \textbf{6.722} & 6.621 
                & \\
\bottomrule
\end{tabular}
\end{table*}

\section{Limitations}
One key constraint is the limited availability of geospatial embedding models specifically tailored to urban environments and based on rich contextual data such as OpenStreetMap (OSM). Moreover, the few existing models are simplistic in nature, often lacking architectural sophistication or the capacity to capture nuanced urban semantics. As a result, the space of existing approaches available for comparative evaluation is currently narrow, which restricts the diversity of baselines and hinders comprehensive benchmarking.

Furthermore, the proposed trajectory tasks have been evaluated in a single resolution, as this resolution offered an optimal trade-off between region precision and the necessity of data augmentation (i.e. the augmentation of missing regions). However, given the higher difficulty level exhibited by the trajectory tasks, it is likely that the different resolutions would facilitate a more extensive evaluation context. 

It is also important to emphasise that the baseline models incorporated within this benchmark are deliberately elementary and essentially function as reference implementations. While these approaches offer consistency and facilitate comparison, they do not reflect the upper bounds of task performance. More sophisticated modelling strategies may be necessary to capture the full complexity of the benchmark tasks.

\section{Future works}
We aim to expand the benchmark by incorporating additional datasets and introducing new task that reflect emerging challenges in geospatial representation learning. The modular design of the accompanying \texttt{SRAI} library allows for seamless integration of such extensions, facilitating community contributions and experimentation. Moreover, we plan to develop a publicly accessible leaderboard that will enable researchers to easily submit and compare results across formulated tasks and datasets. This will foster transparency, reproducibility, and collaborative progress within the geospatial machine learning community. Future work also includes the development of more advanced and task-relevant baseline models to provide stronger reference points and enable more informative evaluation of representation learning methods.
\begin{acks}
Thank you to Kamil Raczycki, Szymon Woźniak and Kacper Leśniara from the Kraina AI team, who supported the technical design of this benchmark. Related Works section was written with the support of deep-research LLM tools. This research was funded in whole or in part by the National Science Centre, Poland grants: Preludium: 2024/53/N/ST6/03296 and Sonata: 2024/55/D/ST6/03082.
\end{acks}

\bibliographystyle{ACM-Reference-Format}
\bibliography{references}

\appendix \label{appendix}

\section{Architecture \& Hyperparameters}\label{appendix:hyper}
All implementations along with exemplary usage notebooks, are publicly available in the \textsc{SRAI} library.

\subsection{Embedders}

All OSM-based embedders compared in the benchmark use a consistent preprocessing pipeline, including the same OpenStreetMap tag filtering strategy \footnote{Following the tag selection procedure from \cite{Wo_niak_2021}, \url{https://github.com/kraina-ai/srai/blob/main/srai/loaders/osm_loaders/filters/hex2vec.py}}. For both GeoVex and ContextualCountEmbedder, we set the spatial neighborhood size to $k=2$ to enable aggregation of local features across adjacent H3 cells. The Hex2Vec embedder was configured with encoder layer sizes of [150, 75, 50] for region-based tasks and [150, 75, 25] for trajectory-based tasks, reflecting the different complexity and granularity requirements of the respective inputs.

\subsection{STRPP \& HPP baseline models}

We use a feedforward regression model with hidden layers [50, 100, 50], ReLU activations, and dropout (p=0.2) after every second layer. The loss function is Smooth L1 Loss, optimized with Adam (lr=0.001). Models were trained for 50 epochs with a batch size of 32.

\subsection{CAP baseline model}

We utilize the same feedforward architecture as in STRPP and HPP, with hidden layers [50,100,50] and Sigmoid activations. The model is trained using L1 Loss and optimized with Adam (lr=0.001) for 50 epochs with a batch size of 32.

\subsection{TTE baseline model}

We use a 2-layer LSTM with hidden size 128 and ReLU activation for travel time regression. The input is a sequence of region embeddings, and the model outputs a single scalar as duration. The model is trained with L1 loss and Adam optimizer (lr=0.001), batch size 32.

\subsection{HMP baseline model}

We employed a two-layer LSTM with hidden size 128 followed by multi-head self-attention and a classifier. The model is trained with a hybrid loss: weighted sum of cross-entropy and log-haversine distance (geo loss weight = 0.7). Optimization uses Adam (lr=0.001), batch size 32, and 10 epochs. During training, ground truth embeddings are iteratively appended using teacher forcing with embedding lookup. During inference, they are switched to predicted cells.

\end{document}